\documentclass[letterpaper, 10 pt, conference]{ieeeconf}  % Comment this line out if you need a4paper

\IEEEoverridecommandlockouts                              % This command is only needed if 
\usepackage{graphics} % for pdf, bitmapped graphics files
\usepackage{epsfig} % for postscript graphics files
\usepackage{mathptmx} % assumes new font selection scheme installed
\usepackage{times} % assumes new font selection scheme installed
\usepackage{cite}
\usepackage{amsmath,amssymb,amsfonts}
\usepackage{textcomp}
\usepackage{xcolor}
\usepackage{booktabs}
\usepackage{algorithm}
\usepackage{algorithmic}
\usepackage{setspace}
\usepackage{ragged2e}
\usepackage{rotating}
\usepackage{subfigure}
\usepackage{multicol}
\usepackage{multirow}
\usepackage{ccaption}
\usepackage{epstopdf}
\usepackage{flushend}
\usepackage{url}
\usepackage{bm}
\usepackage{threeparttable}
\usepackage{array}
\usepackage{makecell}

\title{\LARGE \bf
EMV-LIO: An Efficient Multiple Vision aided LiDAR-Inertial Odometry
}

\author{Bingqi Shen, Yuyin Chen, Fuzhang Han, Shuwei Dai, Rong Xiong, and Yue Wang% <-this % stops a space
\thanks{Bingqi Shen, Fuzhang Han, Rong Xiong and Yue Wang are with the State Key Laboratory of Industrial Control Technology and Institute of Cyber-Systems and Control, Zhejiang University, Hangzhou, China. Yuyin Chen and Shuwei Dai are with Hangzhou Iplus Technology Co., Ltd, Hangzhou, China. Yue Wang is the corresponding author.}%
\thanks{Corresponding author,
        {\tt\small wangyue@iipc.zju.edu.cn}}%
}

\begin{document}

\maketitle
\thispagestyle{empty}
\pagestyle{empty}

%%%%%%%%%%%%%%%%%%%%%%%%%%%%%%%%%%%%%%%%%%%%%%%%%%%%%%%%%%%%%%%%%%%%%%%%%%%%%%%%
\begin{abstract}

To deal with the degeneration caused by the incomplete constraints of single sensor, multi-sensor fusion strategies especially in the LiDAR-vision-inertial fusion area have attracted much interest from both the industry and the research community in recent years. Considering that a monocular camera is vulnerable to the influence of ambient light from a certain direction and fails, which makes the system degrade into a LiDAR-inertial system, multiple cameras are introduced to expand the visual observation so as to improve the accuracy and robustness of the system. Besides, removing LiDAR's noise via range image, setting condition for nearest neighbor search, and replacing kd-Tree with ikd-Tree are also introduced to enhance the efficiency. Based on the above, we propose an Efficient Multiple vision aided LiDAR-inertial odometry system (EMV-LIO), and evaluate its performance on both open datasets and our custom datasets. Experiments show that the algorithm is helpful to improve the accuracy, robustness, and efficiency of the whole system compared with LVI-SAM \cite{b4}. Our implementation is available at \url{https://github.com/BingqiShen/EMV-LIO}. 

\end{abstract}

%%%%%%%%%%%%%%%%%%%%%%%%%%%%%%%%%%%%%%%%%%%%%%%%%%%%%%%%%%%%%%%%%%%%%%%%%%%%%%%%
\section{Introduction}

Recent years have seen great development in the field of application of robots, especially in the domain of navigating uncharted terrains devoid of GPS signals \cite{b1}. To localize the mobile robots in unknown environments, different sensors have been adopted while cameras and LiDARs are the most commonly used among them \cite{b2}. It is well known that LiDAR-based methods can capture the details of the environment from considerable distances, thus acquiring comprehensive geometric information of the surroundings. However, this method often encounters challenges in degenerated environments such as a long tunnel or an open field, since features extracted by LiDAR are almost the same everywhere, rendering it less effective in such scenarios\cite{b3} \cite{b24}. While vision-based methods are especially suitable for place recognition tasks and perform well in texture-rich environments, most of them are extremely sensitive to light changes, fast motion, and initialization.

To solve these problems caused by the incomplete constraints of single sensor, multi-sensor fusion strategies for LiDAR-visual-inertial odometry (LVIO) have attracted much attention from both the industry and the research community in recent years and achieved anti-degenerated state estimation in aforementioned environments since LiDAR-inertial odometry (LIO) contributes precise feature depth to avoid initialization failure of visual-inertial odometry (VIO) while VIO offers an initial pose estimation for LIO. Considering that the measurements of a monocular camera are easily limited by its field of vision (FoV), chances are that the monocular vision aided LiDAR-inertial Odometry system (MoV-LIO) will degrade into LIO or even fail due to the fact that the VIO subsystem provides the wrong initial value of pose for the LIO subsystem. Besides, current LIO systems that can achieve high accuracy are mostly optimization-based such as \cite{b22} \cite{b23}. However, few of them can achieve real-time performance on resource-constrained platforms due to the high computational demands of iterative solving of nonlinear systems. Therefore, it's essential for us to deal with the issue that how to provide a more accurate and robust fusion method under the condition of limited computing resources.

\begin{figure}[t]
    \centering
    \includegraphics[width=0.48\textwidth]{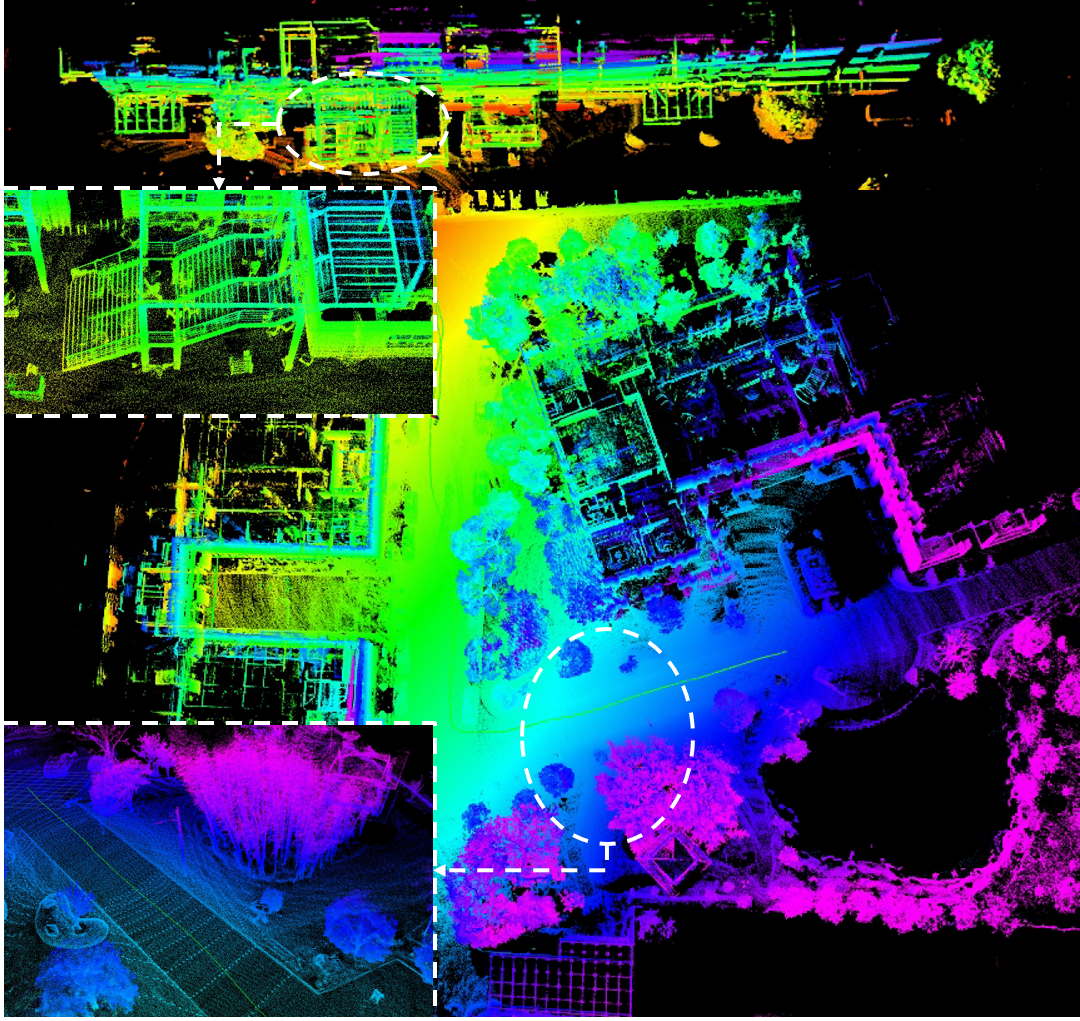} 

    \caption{We use our proposed EMV-LIO to reconstruct a 3D point cloud map of the surroundings of Iplus Technology Co., Ltd. The green path is the computed trajectory and the 3D points are colored by height.}
    \vspace{-0.2cm}
\end{figure}

% Nevertheless, considering that the measurements of a monocular camera are easily limited by its field of vision, chances are that the monocular vision aided LiDAR-inertial system (MV-LIS) will degrade into LiDAR-inertial system (LIS) when facing strong light or a texture-less white wall, where degeneration are still prone to happen. 

In this paper, we propose an Efficient Multiple vision aided LiDAR-inertial odometry system (EMV-LIO) based on LVI-SAM \cite{b4}, introducing multiple cameras in the VIO subsystem to expand the range of visual observation to guarantee the whole system can still maintain the relatively high accuracy in case of the failure of the monocular visual observation. Apart from this, an efficiency-enhanced LVIO system is also introduced to increase the system's efficiency, including removing LiDAR's noise via range image, setting condition for nearest neighbor search, and replacing kd-Tree with ikd-Tree. Experiments show that our proposed method is more robust than LVI-SAM and more accurate than LIO-SAM \cite{b10} facing challenging scenes without affecting efficiency. In summary, the main contributions of this paper are that:

1) Multiple cameras are introduced in the VIO subsystem for more robust and accurate results.

2) An efficiency-enhanced LVIO system is introduced by removing LiDAR's noise via range image, setting condition for nearest neighbor search, and replacing kd-Tree with ikd-Tree to improve the efficiency of the whole system.

3) An Efficient Multiple vision aided LiDAR-inertial odometry system (EMV-LIO) is designed and experiments on both open datasets and our custom datasets are carried out. Experiments show that our system outperforms in robustness and accuracy while maintaining efficiency(see Fig. 1).

The rest of this paper is organized as follows: Section II discusses related work on LiDAR-visual-inertial SLAM, and Section III elaborates on our proposed algorithm. To testify to our analysis, we conduct relevant experiments and the results are presented in Section IV. Ultimately, the whole work is concluded in Section V.

\section{Related Work}
    \subsection{Loosely-coupled LiDAR-visual-inertial Odometry}
    The early research work on LiDAR-visual-inertial fusion was V-LOAM \cite{b5} proposed by the author of LOAM \cite{b6}, Zhang et al, based on their previous research. The issue of robots encountering positioning failures during high-motion scenarios was examined, highlighting its resemblance to the degeneracy challenge stemming from sparse feature data. This observation offers valuable insight for addressing SLAM difficulties in environments characterized by degeneration. In 2019, Wang \cite{b7} combined VINS-mono \cite{b9} and LOAM in a novel approach. Here, VINS-mono functioned as a Visual-Inertial Odometry (VIO) subsystem to provide the initial pose value. Subsequently, the outcome is fed into LOAM to further optimize the pose by minimizing the residual of extracted feature points. In 2020, Khattack introduced CompSLAM, which is a loosely coupled filter combining LOAM, ROVIO \cite{b26}, and kinematic-inertial odometry (TSIF \cite{b27}), thereby demonstrating a comprehensive fusion of these components. More recently, Wang \cite{b25} proposed an advanced SLAM framework that directly fuses visual and LIDAR data. This framework comprises three integral components: a frame-by-frame tracking module, an enhanced sliding window-based refinement module, and a parallel global and local search loop closure detection (PGLS-LCD) module. It ingeniously combines the bag-of-visual-words (BoW) technique with LIDAR iris features to achieve effective location recognition.

    \subsection{Tightly-coupled LiDAR-visual-inertial Odometry}
    Given the fact that the above fusion methods are loosely-coupled, where the LiDAR measurements are not jointly optimized with the visual or inertial measurements. In order to enhance the anti-degeneration ability and robustness of the system, an increasing number of scholars have directed their efforts towards these tightly-coupled approaches. In 2018, Graeter launched LIMO \cite{b8}, pioneering the concept of leveraging LiDAR point depth information to estimate the depth of visual feature points. This innovative approach effectively creates a system similar to RGB-D cameras for the first time. Apart from this, LIC-Fusion \cite{b28} presents a distinctive real-time spatial and temporal multi-sensor calibration, utilizing MSCKF \cite{b29} framework as its foundation. Nonetheless, it is worth mentioning that this work is not publicly available. Inspired by the aforementioned idea, Shan developed LVI-SAM \cite{b4}, which combines LIO-SAM \cite{b10} with VINS-mono. Notably, in scenarios where the LIO subsystem or the VIO subsystem fails, the other system can still remain functional autonomously. This dual-system setup safeguards the overall system's robustness. At the same time, Lin introduced R2LIVE\cite{b12} based on FAST-LIO2 \cite{b11} to fuse the measurements of LiDAR, IMU, and camera within the framework of ES-IKF (Error State Iterative Kalman Filter), which achieves better real-time performance compared with LVI-SAM. Additionally, R2LIVE adopts a comprehensive approach by incorporating all measurements into a factor graph. It optimizes the visual landmarks present within the local sliding window, thereby contributing to the refinement of visual features within the map, and consequently enhancing accuracy. Then in 2021, they launched R3LIVE \cite{b13} on the basis of the above work. This iteration refined the VIO algorithm and rendered the LiDAR point cloud. Similar two submodules, LIO and VIO are also used in FAST-LIVO \cite{b14} and achieve real-time performance at the level of state-of-the-art.

\section{Methodology}
In this section, we describe in detail the proposed EMV-LIO framework. EMV-LIO mainly includes two parts: Multi-camera VIO subsystem and Efficiency-enhanced LVIO system, which will be introduced in the following subsections.

    \subsection{Assumptions and Notations}\label{AA}
    We assume that the intrinsic parameters of each camera are known. The extrinsic parameters between the three kinds of sensors are calibrated, and they are time synchronized. The relationship of different frames is shown in Fig. 2.

    Let us denote the pose of the robot under the World coordinate system at each timestamp $t_k$ as ${\chi}_{k}$ as follows:
    
    \begin{equation}
        {\boldsymbol{\chi}}_{k}=\{ {R_B^O}_k, {p_B^O}_k \} \label{eq}
    \end{equation}

    \begin{flushleft}
    where ${R_B^O}_k$ refer to the 3D rotations from base to odometry at timestamp k while ${p_B^O}_k$ refer to the 3D translations.
    \end{flushleft}

    \begin{figure}[t]
        \centering
        \includegraphics[width=0.48\textwidth]{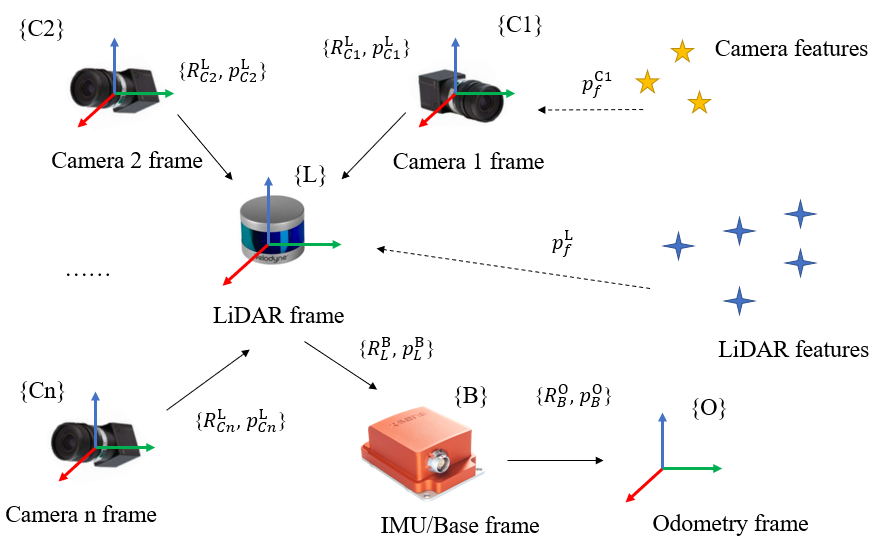} 
    
        \caption{Illustration of each frame.}
        \vspace{-0.2cm}
    \end{figure}

    \subsection{Multi-camera VIO Subsystem} 
    The Multi-camera VIO in our system develops from VINS-mono. All visual features from multiple cameras are detected using a corner detector \cite{b15} first and tracked by the Kanade–Lucas–Tomasi algorithm \cite{b16}. Each feature point is endowed with a unique ID which is managed by establishing a one-way mapping between the feature point set and the camera set. After that, either camera will be selected for initialization by performing Structure from Motion (SfM), including estimating the camera's motion based on epipolar geometry constraint, estimating the depth of feature points by triangulation, and optimizing the camera's pose and the position of feature points by solving the Perspective-n-Point (PnP) problem by Bundle Adjustment (BA). By setting the position of the first frame of the camera $c_o$ as the origin of the reference coordinate system of SfM, we can calculate the pose of the camera for initialization in the IMU coordinate system on the premise of knowing the camera-IMU external parameters ($\bm{R}_C^B, \bm{p}_C^B$).

    \begin{equation}
        \left\{
        \begin{aligned}
        \begin{split}
            \bm{R}_{b_k}^{c_0}&=\bm{R}_{c_k}^{c_0} \cdot  {\bm{R}_C^B} ^{-1}   \\
            s\bm{p}_{b_k}^{c_0}&=s\bm{p}_{c_k}^{c_0} + \bm{R}_{b_k}^{c_0}\bm{p}_C^B
        \end{split}
        \end{aligned}
        \right.
    \end{equation}

    \noindent where $s$ is the unknown scaling parameter of SfM. ($\bm{R}_{b_k}^{c_0}, \bm{p}_{b_k}^{c_0}$) is the relative pose from $b_k$ to $c_0$, and ($\bm{R}_{c_k}^{c_0}, \bm{p}_{c_k}^{c_0}$) is that from $c_k$ to $c_0$. Considering that it is often inaccurate during initialization, the data association between visual features and LiDAR's points will be conducted so that the depth information of the LiDAR's point can be used as the initial value of its corresponding visual feature point for optimization.

    Assuming that there are $n$ cameras in the system, we perform bundle adjustment in a sliding-window which adds the observation residuals of feature points from all cameras into the following cost function:

     % \begin{equation}
        \begin{align}
        % \begin{split}
            &arg min \left\{{\parallel \bm{r}_p - \bm{H}_p \bm{\chi}_k \parallel}^2 + \sum {\parallel \bm{r}_{B}\left( \hat{\bm{z}}_{b_{k+1}}^{b_k}, \bm{\chi}_k\right)\parallel}^2_{\bm{P}_{b_{k+1}}^{b_k}} \right.\nonumber\\ &\left. + \sum \limits_{i=1}^n \sum {\parallel \bm{r}_{C_i}\left( \hat{\bm{z}}_{l}^{c_j}, \bm{\chi}_k\right)\parallel}^2_{\bm{P}_{l}^{c_j}} \right\}
        % \end{split}    
        \end{align}
    % \end{equation}

    \noindent where $\bm{r}_{B}\left( \hat{\bm{z}}_{b_{k+1}}^{b_k}, \bm{\chi}_k\right)$ and $\bm{r}_{C_i}\left( \hat{\bm{z}}_{l}^{c_j}, \bm{\chi}_k\right)$ are residuals for IMU and visual measurements from the $i^{th}$ camera, respectively. $\{\bm{r}_p, \bm{H}_p\}$ is the prior information from marginalization. $\bm{P}_{b_{k+1}}^{b_k}$ is the covariance matrix of IMU from timestamp $k$ to timestamp $k+1$ and $\bm{P}_{l}^{c_j}$ is the covariance matrix of the $l^{th}$ visual feature in the $j^{th}$ frame. All variables above can be deduced from \cite{b9}. The Ceres solver \cite{b21} is used for solving this nonlinear problem.

    \subsection{Efficiency-enhanced LVIO System}
    After obtaining the estimation from VIO as mentioned before, it is feasible for us to utilize it as the initial guess of scan-matching in LIO, which can be solved by minimizing the following cost function \cite{b4}. 
    
    \begin{equation}
        arg min \left \{ \sum \limits_{p_{i, k}^e \in F_k^e} {\bm{d}_{e_k}} + \sum \limits_{p_{j, k}^p \in F_k^p} \bm{d}_{p_k} \right \}
    \end{equation}

    \noindent where $\bm{d}_{e_k}$ and $\bm{d}_{p_k}$ refer to the distance between a feature and its edge or planar patch correspondence at timestamp $k$ relatively. $p_{i, k}^e$ is the $i^{th}$ edge feature and $p_{j, k}^p$ is the $j^{th}$ planar feature while $F_k^e$ and $F_k^p$ are the set of edge features and planar features at timestamp $k$. All of them can be referred from \cite{b10}. To accelerate the process of scan-matching, three tips are proposed as follows:
    
    \begin{itemize}
        \item \textbf{Removing LiDAR's noise via range image:}
         It's well known that the raw data processing including feature extraction and distortion removal is typically the first step in the LiDAR-based SLAM system before further analysis is performed. Considering that the noise in the raw point cloud will decrease the accuracy of odometry while increasing the processing time for the back-end optimization, it is essential to remove the noise at the very beginning of the LIO. In this paper, a method referred from \cite{b17} is adopted. With knowing the vertical scanning angles of a LiDAR, we can project the raw point cloud onto a range image where each valid point is represented by a pixel and the value of it records the Euclidean distance from the point to the LiDAR's origin. Supposing that the horizontal angular resolutions is $\alpha$, clustering is performed assuming that two neighboring points in the horizontal or vertical direction belong to the same object if the angle $\gamma$ between their connected line and the laser beam is above a certain threshold $\theta$. $\gamma$ can be calculated by applying trigonometric equations.

         \begin{equation}
            \gamma = artan \frac{d_{2}sin\alpha}{d_{1}-d_{2}cos\alpha}
         \end{equation}
         
         where $d_1$ and $d_2$ are the depths of two adjacent points belonging to the same ring which can be obtained from the range image. Then we discard small clusters since noise may be contained and it will offer unreliable constraints in optimization.

        \item \textbf{Setting condition for nearest neighbor search:}
        Most of the LiDAR-based SLAM perform Nearest Neighbor Search (NNS) to find the nearest neighbor point in the previous frame or local map for each point in the current scan during each iteration, which is the most time-consuming step during the whole optimization process. Considering that when the optimization process is about to end, chances are that the result of the current NNS is the same as the result of the previous iteration since the optimization result tends to converge, which leads to unnecessary searches. In this paper, we propose to skip this round of search and adopt the result of the previous round of search once $\Delta \boldsymbol{\chi}_{ki} < {\alpha} \Delta \boldsymbol{\chi}_{thres}$, where $\Delta \boldsymbol{\chi}_{ki}$ is the incremental iteration result of the $i^{th}$ round at timestamp $t_k$. $\Delta \boldsymbol{\chi}_{thres}$ is the convergence threshold of iteration and $\alpha$ is a ratio which can be set as 2.
        
        \item \textbf{Replacing kd-Tree with ikd-Tree:}
        The data structure used for NNS is usually kd-Tree. However, it does not support incremental updates, which means that after building kd-Tree for the current frame point cloud, new points can be added to the frame point cloud only by building a new kd-Tree, which adds additional computing overhead. Therefore, we adopt incremental kd-Tree (ikd-Tree) proposed by Fast-LIO2 \cite{b11}, which can incrementally insert and delete points and be automatically re-balanced by partial re-building which enables efficient NNS in later stages.
        
    \end{itemize}

\section{Experiments}
    % \subsection{System integration and experimental setting}
    For the purpose of ensuring the authenticity and objectivity of the experimental results, we conducted experiments on both open datasets and our custom datasets. Open datasets we choose for accuracy and efficiency test are $Quad$-$Medium$ and $Quad$-$Hard$ sequences of the Newer College Multi-camera Dataset (NCD-Multi) \cite{b18} which uses an AlphaSense multi-camera sensor with synchronized IMU and Ouster LiDAR. Besides, we name $AS\_C0$ as the front camera, $AS\_C3$ as the right camera, and $AS\_C4$ as the left camera as shown in Fig. 3(a) for convenience. Our custom datasets for efficiency tests are collected by three cameras, a LiDAR and an IMU as shown in Fig. 3(b). The hardware and overview of the above datasets are listed in Table 1, from which we can conclude that each dataset contains fast motion scenes and there are also some challenging scenes like strong light or being occluded by a wall as shown in Fig. 4. Each experiment was conducted using a desktop PC with an Intel Core i7-6700 with 64 GB RAM.

    \begin{figure}[t]
    	\centering
    	\subfigure[]{
                \begin{minipage}[t]{0.45\linewidth} 
    		      \includegraphics[width=\textwidth]{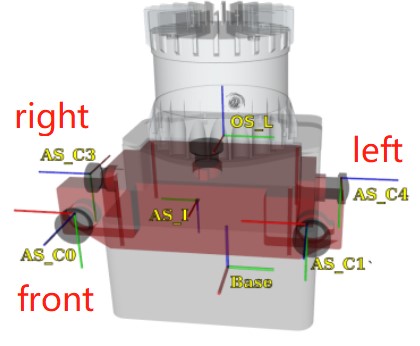} 
    		\end{minipage}
    	}
            \subfigure[]{
                \begin{minipage}[t]{0.45\linewidth} 
    		      \includegraphics[width=\textwidth]{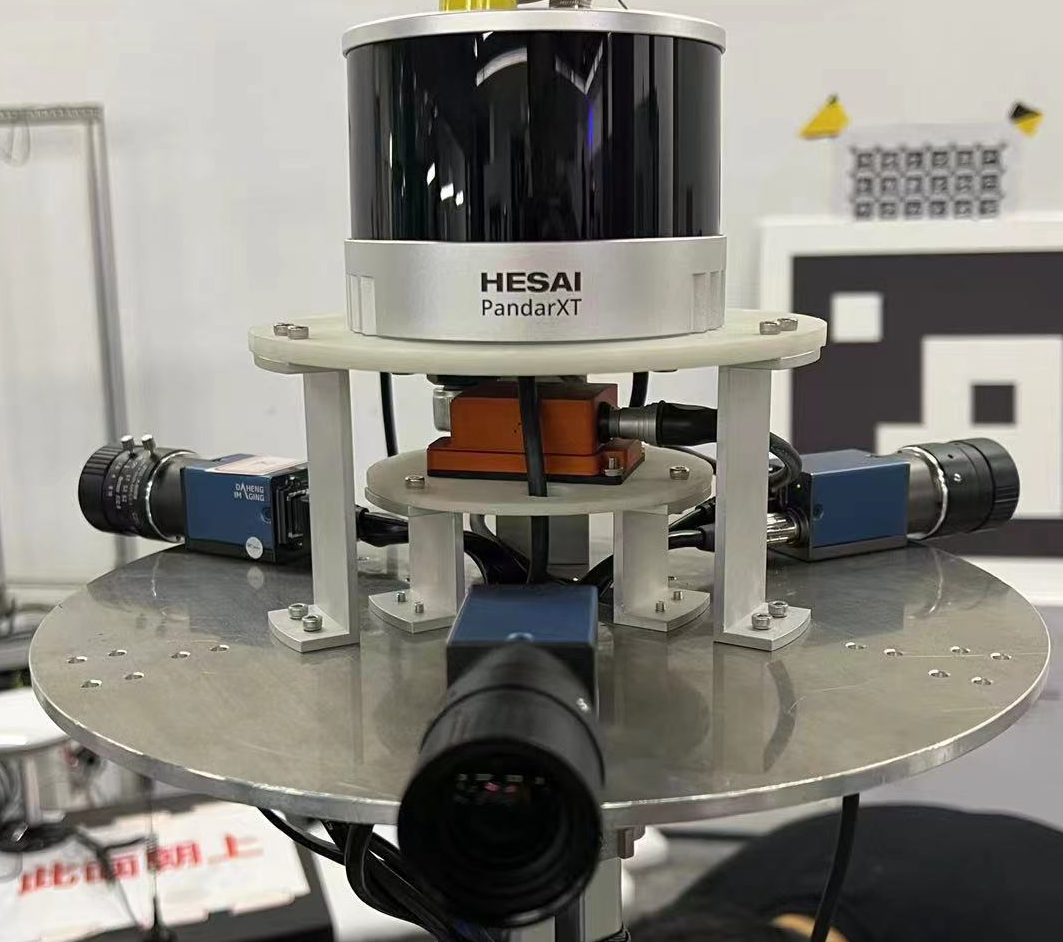} 
    		\end{minipage}
    	}

    	\caption{(a) Sensors used in NCD-Multi with their corresponding coordinate systems. (b) Sensors used for collecting our custom dataset including three cameras, a LiDAR, and an IMU. All of them are synchronized.}
    	\vspace{-0.3cm}
    \end{figure}

    \begin{table}[htbp] 
        \scriptsize
        \caption{\label{tab:test}Hardware and overview of datasets used for experiments, including max acceleration (for short: max acc., unit: $m/s^2$) and max angular velocity (for short max angv., unit: rad/s) given by IMU.} 
        \centering
        % \subtable[]{
            \begin{tabular}{ccccc} %需要5列
            \toprule 
                Dataset & LiDAR & Camera & IMU & Max acc/angv.\\ 
                \midrule 
                $Quad$-$Medium$ & OS0-128 & Alphasense & BMI085 & 13.03/1.54 \\ 
                $Quad$-$Hard$ & OS0-128 & Alphasense & BMI085 & 12.64/2.34 \\
                Custom & Pandar XT-32 & MER2-202 & MTI-300 & 20.21/0.58\\
            \bottomrule 
        \end{tabular} 
        % }
    \end{table}

    \begin{figure}[t]
    	\centering
    	\subfigure[Strong light]{
                \begin{minipage}[t]{0.45\linewidth} 
    		      \includegraphics[width=\textwidth]{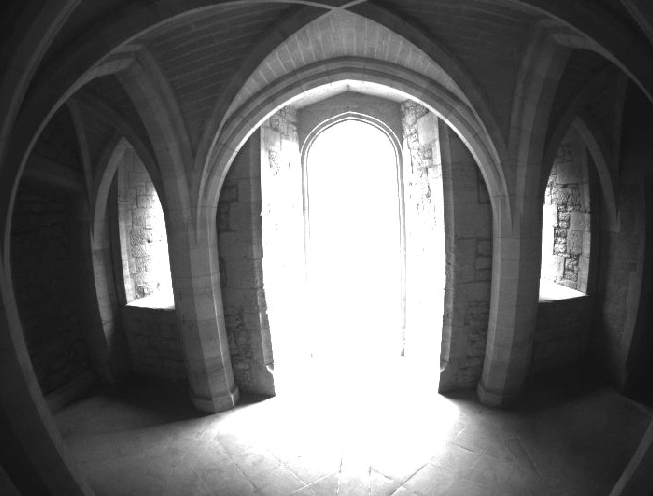} 
    		\end{minipage}
    	}
            \subfigure[Occluded by a wall]{
                \begin{minipage}[t]{0.45\linewidth} 
    		      \includegraphics[width=\textwidth]{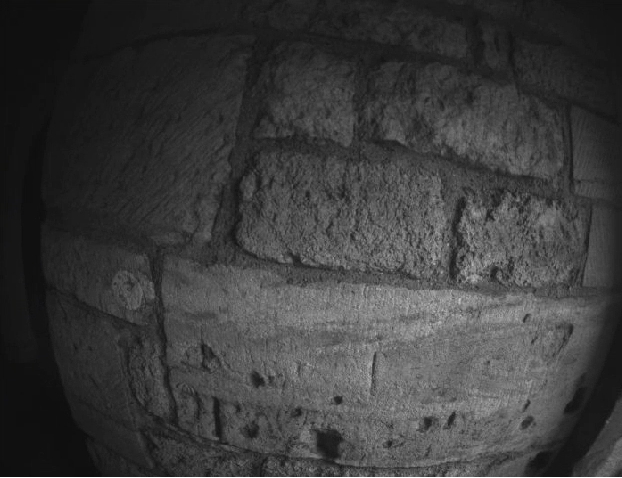} 
    		\end{minipage}
    	}

    	\caption{Challenging scenes of the Newer College Multi-camera Dataset}
    	\vspace{-0.3cm}
    \end{figure}

    \subsection{Accuracy Test}
    \begin{table}[htbp] 
        \scriptsize
        \caption{\label{tab:test}RPE Information [m] Based on Different Methods} 
        \centering
        \begin{threeparttable}
        \subtable[$Quad$-$Medium$ sequence]{
            \begin{tabular}{ccccc} 
                \toprule 
                Method & RMSE & MEAN & MAX & MIN \\ 
                \midrule 
                \textbf{LIO-SAM} & 0.0975 & 0.0857 & \textbf{0.3829} & 0.0047 \\
                \textbf{LVI-SAM (front)} & - & - & - & - \\ 
                \textbf{LVI-SAM (right)} & - & - & - & - \\
                \textbf{LVI-SAM (left)} & 0.0978 & \textbf{0.0854} & 0.4123 & 0.0075 \\
                \textbf{EMV-LIO (front + right)} & 0.0982 & 0.0864 & 0.4002 & \textbf{0.0046} \\
                \textbf{EMV-LIO (front + left)} & \textbf{0.0969} & 0.0861 & 0.4610 & 0.0073 \\
                \textbf{EMV-LIO (right + left)} & 0.0974 & 0.0866 & 0.4857 & 0.0055 \\
                \textbf{EMV-LIO (front + right + left)} & 0.0976 & 0.0868 & 0.4602 & 0.0081 \\
                \bottomrule 
            \end{tabular} 
        }
        
        \subtable[$Quad$-$Hard$ sequence]{
            \begin{tabular}{ccccc} 
                \toprule 
                Method & RMSE & MEAN & MAX & MIN \\ 
                \midrule 
                \textbf{LIO-SAM} & 0.2140 & 0.1862 & 0.7270 & \textbf{0.0043} \\
                \textbf{LVI-SAM (front)} & - & - & - & - \\ 
                \textbf{LVI-SAM (right)} & - & - & - & - \\
                \textbf{LVI-SAM (left)} & 0.3586 & 0.2325 & 1.5116 & 0.0168 \\
                \textbf{EMV-LIO (front + right)} & 0.1416 & 0.1294 & 0.4448 & 0.0113 \\
                \textbf{EMV-LIO (front + left)} & 0.1411 & 0.1287 & 0.4503 & 0.0083 \\
                \textbf{EMV-LIO (right + left)} & 0.1426 & 0.1296 & 0.4175 & 0.0095 \\
                \textbf{EMV-LIO (front + right + left)} & \textbf{0.1362} & \textbf{0.1239} & \textbf{0.4000} & 0.0107 \\
                \bottomrule 
            \end{tabular} 
        }

        \begin{tablenotes}
            \footnotesize
            \item[$-$] represents failure of odometry.
        \end{tablenotes}
        \end{threeparttable}
    \end{table}

    Experiments on LIO-SAM, LVI-SAM using different cameras, and our proposed EMV-LIO using different combinations of cameras were conducted and Relative Pose Errors (RPE) described in \cite{b19} were adopted to evaluate the performances. The RPE information estimated by different methods is summarized in Table \uppercase\expandafter{\romannumeral2}. Since there are many challenging scenes in the two sequences like fast motion or being occluded by the wall, the failure of VIO is likely to occur in LVI-SAM depending on a monocular camera so that a wrong initial guess may be provided for LIO, leading to a worse result compared with LIO-SAM. Nevertheless, for EMV-LIO using multiple cameras proposed in this paper, when one of the above cameras fails due to inadequate trackable features caused by challenging scenes, the feature points tracked by the other cameras are relatively stable at this time, which can provide more accurate visual-inertial odometry result for the LiDAR-inertial odometry as a priori, thus improving the robustness of the overall system, which is a vivid annotation of \textit{Do not put all your eggs into one basket}. In addition, the introduction of more visual re-projection error items during optimization can bring a certain degree of accuracy improvement.

    \subsection{Efficiency Test}

    \begin{figure}[t]
        \centering
            \subfigure[Point cloud before noise removal]{
                \includegraphics[width=0.4\textwidth]{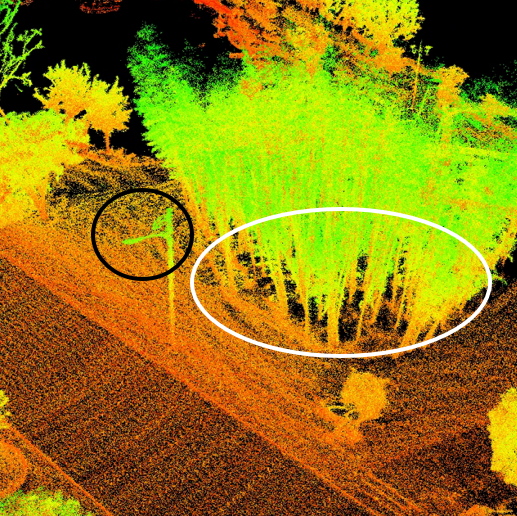} 
                \centering
            }
                \subfigure[Point cloud after noise removal]{
                \includegraphics[width=0.4\textwidth]{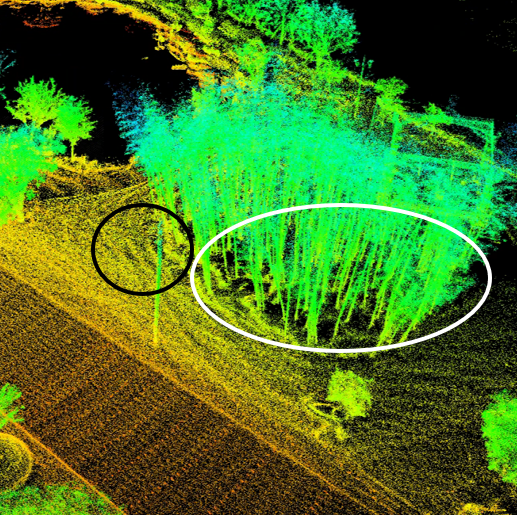} 
                \centering
            }
    
        \caption{Comparison between before and after noise removal}
        \vspace{-0.3cm}
    \end{figure}

    \begin{table*}[!htbp] 
        \scriptsize
        \caption{\label{tab:test}Efficiency Information Based on Different Methods} 
        \centering
        
            \begin{tabular}{c c c c c c c c c c c c c c} 
                \toprule
                \multicolumn{3}{c}{\multirow{2}{*}{Method}} & \multicolumn{3}{c}{\textbf{LIO-Prehandle}} & \multicolumn{5}{c}{\textbf{LIO-Optimization}} &
                \multicolumn{2}{c}{\textbf{VIO}} \\
                \cmidrule(lr){4-6} \cmidrule(lr){7-11} \cmidrule(lr){12-13}
                \multicolumn{3}{c}{}& Number of point & Total time (ms) & CPU (\%) && NNS (ms) & Extracting KF & Total time (ms) & CPU (\%) & Total time (ms) & CPU(\%)\\  %有n个&，就表示该行有n+1列
                \hline %绘制一条水平横线
                \multicolumn{3}{c}{\textbf{Method A}} & 5817.8 & $\textbf{29.6}$ & $\textbf{60.5}$ &  & 22.3 & 45.8 & 80.1 & 135.2 & 61.3 & 238.4\\   % 占两列，列名为A；后面陆续跟着数字
                \multicolumn{3}{c}{\textbf{Method B}} & 4742.7 & 41.6 & 66.1 &  & 19.8 & 32.4 & 54.2 & 117.2 & 61.3 & 238.4 \\
                \multicolumn{3}{c}{\textbf{Method C}} & 4742.7 & 41.6 & 66.1 &  & 17.8 & 32.3 & 52.2 & 117.3 & 61.3 & 238.4 \\
                \multicolumn{3}{c}{\textbf{Method D}}& $\textbf{4742.7}$ & 41.6 & 66.1 &  & $\textbf{15.1}$ & $\textbf{0.0}$ & $\textbf{17.9}$ & $\textbf{83.7}$ & $\textbf{61.3}$ & $\textbf{238.4}$ \\
                \bottomrule 
            \end{tabular} 

            % \begin{tabular}{c c c c c c c c c c c c c c c} 
            %     \toprule
            %     \multicolumn{3}{c}{\multirow{2}{*}{Method}} & \multicolumn{2}{c}{\textbf{VIO}} & \multicolumn{3}{c}{\textbf{LIO-Prehandle}} & \multicolumn{5}{c}{\textbf{LIO-Optimization}} \\
            %     \cmidrule(lr){4-5} \cmidrule(lr){6-8} \cmidrule(lr){9-14}
            %     \multicolumn{3}{c}{}& Total time (ms) & CPU(\%) & Number of point & Total time (ms) & CPU (\%) && NNS (ms) & Average count of NNS & Extract key frames (ms) & Total time (ms) & CPU (\%)\\  %有n个&，就表示该行有n+1列
            %     \hline %绘制一条水平横线
            %     \multicolumn{3}{c}{\textbf{Method A}}& & & 5817.82 & $\textbf{29.60}$ & $\textbf{60.5}$ &  & 8.37 & 2.66 & 45.83 & 80.12 & 135.2\\   % 占两列，列名为A；后面陆续跟着数字
            %     \multicolumn{3}{c}{\textbf{Method B}}& & & 4742.74 & 41.62 & 66.1 &  & 8.35 & 2.37 & 32.41 & 54.21 & 117.2\\
            %     \multicolumn{3}{c}{\textbf{Method C}}& & & 4742.74 & 41.62 & 66.1 &  & 8.61 & $\textbf{2.07}$ & 32.36 & 52.19 & 117.3\\
            %     \multicolumn{3}{c}{\textbf{Method D}}& & & $\textbf{4742.74}$ & 41.62 & 66.1 &  & $\textbf{6.82}$ & 2.21 & $\textbf{0.0}$ & $\textbf{17.93}$ & $\textbf{83.7}$\\
            %     \bottomrule 
            % \end{tabular} 
        % }
        % \end{center}
        
    \end{table*}

    We divide the LIO subsystem into two main modules, which are the prehandle module and the optimization module. In order to evaluate the efficiency performance, we report the number of processing points in the prehandle module, the time usage of NNS, the time usage of extracting keyframes in the optimization module, and the total time usage and the CPU utilization percentage of the two modules in the LIO subsystem while the total time usage and the CPU utilization percentage of the VIO subsystem are also recorded. For the sake of convenience, different methods will be defined as follows:
    
    \textbf{Method A}: The original LVI-SAM \cite{b4}.
    
    \textbf{Method B}: The method with the addition of removing LiDAR’s noise via range image based on \textbf{Method A}
    
    \textbf{Method C}: The method with the addition of setting condition for nearest neighbor search based on \textbf{Method B}
    
    \textbf{Method D}: The method with the addition of replacing  based on \textbf{Method C}.

    \begin{table}[htbp] 
        \scriptsize
        \caption{\label{tab:test}The relative odometry frame average rate (for short: ravg. rate) Based on Different Methods} 
        \centering
        \setlength{\tabcolsep}{1.5mm}
        \begin{threeparttable}
        \subtable[$Quad$-$Medium$ sequence]{
            \begin{tabular}{cccccc} 
                \toprule 
                Method & \textbf{LIO-SAM} & \textbf{Method A} & \textbf{Method B} & \textbf{Method C} & \textbf{Method D} \\ 
                \midrule 
                Ravg. rate & 0.0\% & -7.6\% & -4.3\% & -3.2\% & \textbf{+2.1\%}\\
                \bottomrule 
            \end{tabular} 
        }
        
        \subtable[$Quad$-$Hard$ sequence]{
            \begin{tabular}{cccccc} 
                \toprule 
                Method & \textbf{LIO-SAM} & \textbf{Method A} & \textbf{Method B} & \textbf{Method C} & \textbf{Method D} \\ 
                \midrule 
                Ravg. rate & \textbf{0.0\%} & -9.7\% & -7.5\% & -6.4\% & -0.2\%\\
                \bottomrule 
            \end{tabular} 
        }

        \end{threeparttable}
    \end{table}

    The efficiency information based on different methods is summarized in Table \uppercase\expandafter{\romannumeral3}. The result of \textbf{Method A} and \textbf{Method B} indicates that removing LiDAR's noise via range image can reduce the number of processing points at the cost of increased time consumption in prehandle step, which can decrease the time consumption in optimization. As shown in Fig. 5 circled by white ovals, the point cloud of the bamboo grove after noise removal is obviously clearer than that of before. Nevertheless, it tends to underperform in reconstructing objects at high places circled by black ovals for instance, since their values in range image change greatly and will be treated as different kinds of objects during clustering and then filtered out. The introduction of setting condition for nearest neighbor search can reduce the search times to help improve the efficiency of the whole system, which can be concluded from the comparison between \textbf{Method B} and \textbf{Method C}. Besides, replacing kd-Tree with ikd-Tree can significantly decrease the CPU utilization percentage and the optimization time consumption. Since the emphasis is only put on the efficiency improvement of the LIO subsystem, the total time usage and the CPU utilization percentage of the VIO subsystem remain the same throughout the experiment. 

    The relative odometry frame average rate based on different methods is summarized in Table \uppercase\expandafter{\romannumeral4}. We set the odometry frame average rate of \textbf{LIO-SAM} as the baseline and report the relative values compared with it. The result of \textbf{LIO-SAM} and \textbf{Method A} shows that the introduction of visual observation can reduce the odometry frame rate intuitively. However, our proposed method can help compensate for the reduction and even perform better than \textbf{LIO-SAM} which does not take visual observation into account.

\section{Conclusions}

Given the fact that the monocular vision aided LiDAR-inertial odometry system (MoV-LIO) is easy to degrade into LiDAR-inertial odometry system (LIO) when facing strong light or a texture-less white wall, where degeneration is still prone to happen, we propose an Efficient Multiple vision aided LiDAR-inertial odometry system (EMV-LIO) based on LVI-SAM \cite{b4}, which introduces visual measurements from multiple cameras to assist odometry. Besides, an efficiency-enhanced LVIO system including removing LiDAR's noise via range image, setting condition for nearest neighbor search, and replacing kd-Tree with ikd-Tree are also presented in this paper. To testify to our proposed method, experiments on accuracy and efficiency were conducted, and the final results show that our proposed method can help improve the accuracy and efficiency of the whole system.

There are several directions for future work. Since we still initialize the VIO subsystem via a monocular camera, chances are initialization via multiple cameras will be taken into account. Besides, Table \uppercase\expandafter{\romannumeral3} shows that the bottleneck of efficiency improvement of our proposed system now is the VIO subsystem, on which will be put more emphasis. Apart from this, some methods such as changing the way of point cloud downsampling or replacing kd-Tree with IVox \cite{b20} can be applied to further improve the efficiency of the whole system.

\addtolength{\textheight}{-12cm}   % This command serves to balance the column lengths

%%%%%%%%%%%%%%%%%%%%%%%%%%%%%%%%%%%%%%%%%%%%%%%%%%%%%%%%%%%%%%%%%%%%%%%%%%%%%%%%

%%%%%%%%%%%%%%%%%%%%%%%%%%%%%%%%%%%%%%%%%%%%%%%%%%%%%%%%%%%%%%%%%%%%%%%%%%%%%%%%

%%%%%%%%%%%%%%%%%%%%%%%%%%%%%%%%%%%%%%%%%%%%%%%%%%%%%%%%%%%%%%%%%%%%%%%%%%%%%%%%

%%%%%%%%%%%%%%%%%%%%%%%%%%%%%%%%%%%%%%%%%%%%%%%%%%%%%%%%%%%%%%%%%%%%%%%%%%%%%%%%

\end{document}